# A high performance approach to detecting small targets in long range low quality infrared videos

Chiman Kwan[1] • Bence Budavari[1]

**Abstract** Since targets are small in long range infrared (IR) videos, it is challenging to accurately detect targets in those videos. In this paper, we propose a high performance approach to detecting small targets in long range and low quality infrared videos. Our approach consists of a video resolution enhancement module, a proven small target detector based on local intensity and gradient (LIG), a connected component (CC) analysis module, and a track association module to connect detections from multiple frames. Extensive experiments using actual mid-wave infrared (MWIR) videos in ranges between 3500 m and 5000 m from a benchmark dataset clearly demonstrated the efficacy of the proposed approach.

## 1 Introduction

Infrared (IR) videos in ground based imagers contain a lot of background clutter and flickering noise due to air turbulence, etc. Moreover, the target size is quite small and hence it is challenging to detect small targets from a long distance.

Small target detection for infrared images has been active in recent years [1]-[6]. Chen *et al.* [1] proposed to detect small IR targets by using local contrast measure (LCM), which is time-consuming and sometimes enhances both targets and clutters. To improve the performance of LCM, Wei *et al.* [2] introduced a multiscale patch-based contrast measure (MPCM). Gao *et al.* [3] developed an infrared patch-image (IPI) model to convert small target detection to an optimization problem. Zhang *et al.* [4] improved the performance of the IPI via non-convex rank approximation minimization (NRAM). Zhang *et al.* [5] proposed to detect small IR targets based on local intensity and gradient (LIG) properties, which has good performance and relatively low computational complexity. Recently, Chen et al. [6] proposed a new and real-time approach for detecting small targets with sky background.

Parallel to the above small target detection activities, there are some conventional target tracking methods [7][8]. Furthermore, some target detection and classification schemes using deep learning algorithms (You Only Look Once (YOLO)) for larger objects in short-range infrared videos have been proposed in the literature [9]-[27]. Since YOLO uses texture information to help the detection, the use of YOLO is not very effective for long range videos in which the targets are too small to have any discernible textures. Some of these new algorithms incorporated compressive measurements directly for detection and classification. Real-time issues have been discussed in [27].

In this paper, we summarize the investigation of an integrated approach to enhancing the performance of small target detection in long range infrared videos. Our approach contains several modules. First, in order to improve the resolution of the low resolution infrared videos, we propose to utilize proven video super-resolution enhancement algorithms to improve the spatial resolution of the video. Three algorithms [28]-[30] were compared and one method was proven to improve the target detection performance. Second, we propose to incorporate a proven target detector known as LIG [5] into our framework. LIG has shown good performance and computational speed in our recent paper [6]. Some customizations of LIG with respect to adaptive threshold selection have been carried out in this paper. Third, even with careful and robust threshold selection in LIG, there are still false positives in the detection results. Dilation within the connected component (CC) analysis module has been used to merge nearby detections together, which in turn mitigates the false positives. Through some rule analysis, the false positives have been further reduced. Finally, in order to further improve the overall detection performance, a simple and fast target association algorithm has been included in our framework. The method is known as Simple Online

✉ Chiman Kwan
chiman.kwan@arllc.net

[1] Applied Research LLC, Rockville Maryland, USA

and Realtime Tracking (SORT) [31], which further enhances the target detection performance.

Our contributions are as follows:

- An integrated framework for target detection in long range videos
  We propose an integrated, flexible, and modular framework comprising video super-resolution, small target detection, connected component (CC) analysis, and target track association.
- Unsupervised target detection approach
  For long range videos (3500m and beyond), the target size is very small. Our proposed framework, unlike YOLO, does not require any training.
- Target detection enhancement using super-resolution images
  The objective here is to investigate whether super-resolution videos will help the small target detection performance. If super-resolution videos do help the detection, we would also like to quantify the detection performance using some well-known metrics such as precision, recall, and F1 score. Three super-resolution approaches were investigated: one conventional and two deep learning. One of them was observed to yield improved target detection results.
- Extensive experiments using long range infrared videos in a benchmark dataset known as SENSIAC [32] containing videos from 3500 m to 5000 m away clearly demonstrated the efficacy of our proposed framework.

The rest of this paper is organized as follows. In Section 2, we will present the technical approach. Section 3 summarizes the experimental results using long range videos from a benchmark dataset. Finally, Section 4 concludes the paper with some remarks and future direction directions.

## 2 Technical approach

Our proposed approach consists of several modules. First, since the long range videos have low resolution, we propose to apply state-of-the-art video super-resolution algorithms to enhance the video resolution. The objective is to investigate how much performance gain one can achieve with video-super-resolution. Second, an unsupervised small target detection using low intensity and gradient (LIG) [5] algorithm is applied to each frame for small target detection. Some improvements over the existing LIG method have been introduced by us. Third, since the LIG detection results may have some scattered false positives, we propose to apply connected component (CC) analysis to group neighboring pixels into clusters. Some empirical rules were formulated in this module. Finally, we propose a fast target associate algorithm known as Simple Online and Realtime Tracking (SORT) [31] to form target tracks. This step turns out to further improve the detection results.

### 2.1 Video Super-Resolution Algorithms

For long range videos, the target size is quite small. Improving the resolution of the videos may help the target detection performance. In this research, we investigated three image resolution enhancement algorithms, which are briefly explained below.

**Bicubic**
Bicubic interpolation is a single frame super-resolution method. Pixels in the original frames are interpolated using 16 neighbors. It has been widely used in many practical applications due to its simplicity and computational efficiency.

**Dynamic Upsampling Filter (DUF) Algorithm for Video Super-Resolution**
Video Super Resolution using Dynamic Upsampling Filter (VSR-DUF) a recent deep learning super resolution method developed by researchers at Yonsei University [28]. This particular model is able to utilize temporal information when generating a high resolution frame. VSR-DUF for instance uses frames before and after the current one in order to generate a single super resolution frame. The number of frames used before or after a given frame is known as the temporal radius. In our studies, we used a radius of 7. This is advantageous over single image super resolution methods since there are more relevant frames to draw information from.

**Zoom Slow-Motion (ZSM) Algorithm for Video Super-Resolution**
The Zooming-Slow Motion (ZSM) is a recent state of the art deep learning video super resolution method [29]. The method was developed by researchers from Purdue University, Northeastern University, and University of Rochester in 2020. The ZSM not only improves the resolution of the frames within a video but also improves the frame rate of the input videos (although the additional frames are currently not being utilized within our workflow).

ZSM can be broken down into three key components: feature temporal interpolation network, a deformable ConvLSTM, and a deep construction network [30]. The feature temporal interpolation network is used to interpolate missing temporal information between the input low resolution frames. Next, the deformable ConvLSTM is used to align and aggregate the temporal information together. Lastly, the deep construction network predicts and generates the super resolution upsampled video frames.

The visual performance of the various super-resolution methods are shown in Fig. 1. It can be seen that ZSM method (two times and four times) yielded better results that DUF and bicubic methods. The edges around the bright spots are much sharper in the ZSM cases compared to the others.

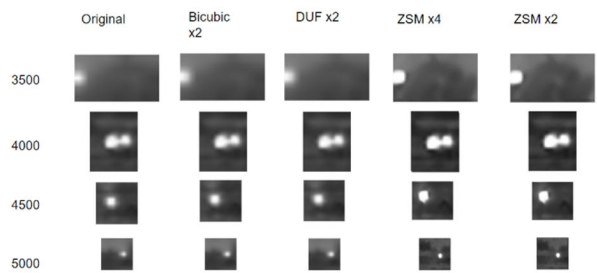

Fig. 1. Comparison of the image qualities of different video super-resolution algorithms. Videos from 3500 m to 5000 m were used in our comparisons.

## 2.2 LIG principle and its improvements

Since the detection results of YOLO at the longer ranges (3500m and above) were not as high as we would have liked [12], we investigated a traditional unsupervised small target detection method to see how it would perform on the long range videos. The algorithm of choice for this study was a local intensity gradient (LIG) based target detector [5], specifically designed for infrared images. The LIG is relatively fast than other algorithms and is very robust to background clutter. The algorithm scans through the input image using a sliding window, whose size depends on the input image resolution. For each window, the local intensity and gradient values are computed separately. Then, those values are multiplied to form an intensity-gradient (IG) map. An adaptive threshold is then used to segment the IG map and then the binarized image will reveal the target.

A major advantage of these unsupervised algorithms is that they require no training so there is no need to worry about customizing training data, which is the case with YOLO. A disadvantage of the LIG algorithm is that it is quite slow, taking roughly 30s per frame of size 640 x 480 pixels.

There are several adjustments we made to the LIG algorithm to make it more suitable for the SENSIAC infrared dataset. First of all, we used differing patch sizes for the different resolution frames. For the original resolution frames we used a patch size of 7x7 and for the 2x upsampled frames, we used a larger patch size of 19x19. Second, we adjusted the way in which the adaptable threshold $T$ is calculated. In [5], the authors used the mean value of all non-zero pixels. For our dataset, this calculation produced a very small value due to the overwhelming amount of very low non-zero pixels. The left image in Fig. 2 highlight the significant role that the threshold plays for this algorithm. Third, we have implemented ways of speeding up the algorithm, such as incorporating multithreading within the script. We were able to speed up the computational time by close to three times.

Here, we would like to illustrate the importance of adaptive thresholding. For the example in Fig. 2, the mean value was 0.008. Using this threshold value for binarization as suggested by [5], we observe that roughly half the non-zero pixels would be considered as detections, as seen on the left hand image of Fig. 2. This originally resulted in hundreds of false positives in the frames. So instead of using the mean of non-zero pixels in the LIG processed frame, we used the mean of the top 0.01% of pixels. A higher threshold is essential for eliminating false positives, as can be seen in the image on the right of Fig. 2.

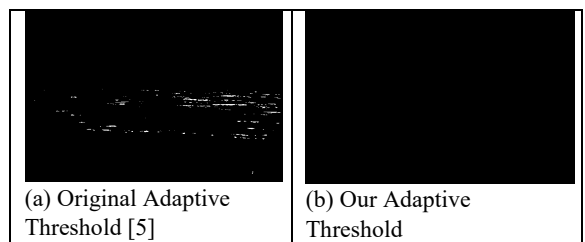

(a) Original Adaptive Threshold [5]    (b) Our Adaptive Threshold

Fig. 2. Segmentation results using two different adaptive thresholds.

## 2.3 CC analysis

We then perform connected component analysis on the segmented binarized image to find groupings of moving pixels between frames. Dilation of the binarized image can be used prior to the analysis in order to merge nearby pixels together. Moreover, the connected components are fed into a "Rule Analysis" block where there are several rule checks to determine whether the connected component is a valid detection or not. These rules involve checking if the area of the connected component is reasonable as well as comparing the max intensity of pixels between the connected components. If the area is over 1 pixel and less than 100 pixels it is valid. Out of the remaining connected components the one with the pixel with the highest intensity is then chosen as the target.

### 2.4 Target Association Using SORT

Another tool we utilized is the SORT tracking association. SORT utilizes motion information as well as memory of past frames to associate targets from one frame to another. Each detected object within a frame is modelled as follows [31]:

$$\mathbf{x} = [u, v, s, r, \dot{u}, \dot{v}, \dot{s}]^T,$$

where $u$ and $v$ represent the horizontal and vertical pixel location of the object center, while $s$ and $r$ indicate the scale and aspect ratio of the object. These object states are then compared across frames to determine whether any of the states are correlated to any previous state.

## 3 Experimental results

### 3.1 Videos

We used four videos in the SENSIAC dataset [32]. There are MWIR daytime and night-time videos ranging from 1000 m to 5000 m in 500 m increments. We selected four daytime MWIR videos in the following ranges: 3500 m, 4000 m, 4500 m, and 5000 m. Fig. 3 to Fig. 6 show several frames from each video. It can be seen that the vehicles (small bright spots) are hard to see and are quite small in size.

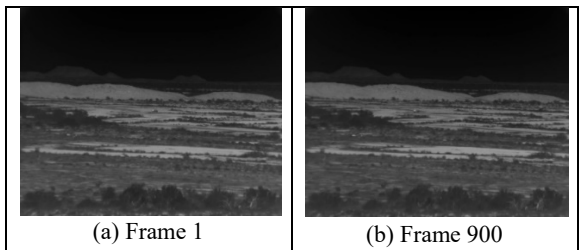
(a) Frame 1      (b) Frame 900
Fig. 3. Frames from the 3500 m video.

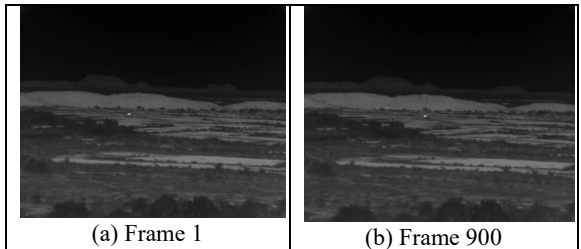
(a) Frame 1      (b) Frame 900
Fig. 4. Frames from the 4000 m video.

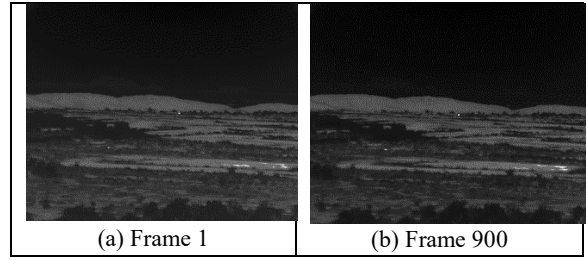
(a) Frame 1      (b) Frame 900
Fig. 5. 4500 m daytime videos.

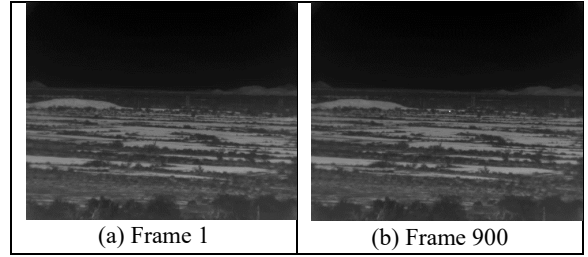
(a) Frame 1      (b) Frame 900
Fig. 6. Frames from the 5000 m video.

### 3.2 Performance metrics

A correct detection or true positive (TP) occurs if the binarized LIG result is within a certain threshold of the centroid of the ground truth bounding box. Otherwise, the detected object is regarded as a false positive (FP). Based on the correct detection and false positives counts, we can further generate precision, recall, and F1 metrics. The precision (P), recall (R), and F1 are defined as

$$P = \frac{TP}{TP+FP} \qquad (1)$$

$$R = \frac{TP}{TP+Missed\ detections} \qquad (2)$$

$$F1 = \frac{2 \times P \times R}{P+R} \qquad (3)$$

### 3.3 Performance of Proposed Approach without SORT

The proposed workflow without SORT is shown in Fig. 7. The VSR and LIG steps were mentioned earlier. The CC analysis involves several steps. First, it binarizes the LIG map with the adaptive threshold, meaning all values below the threshold are 0 and all those above are 1. Once this is done, dilation is performed on the resulting binarized image. The structuring element used is a square and its size depends on the resolution of the images. At the original scale, we used a square with length 5 and at any other resolution we used a square with length 10. The dilation is crucial within the workflow because it merges groups of isolated pixels that correspond to the same target. In other words, without dilation, there would be significantly more false positives throughout the videos.

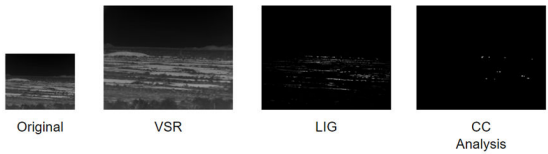

Fig. 7. Illustration of the workflow by combining ZSM and LIG. The video super-resolution (VSR) is done with ZSM.

For our studies, we utilized the pretrained models provided by the authors of DUF and ZSM. The ZSM was trained on the large Vimeo-Septuplet dataset, which includes approximately 90,000 7-frame sequences from a variety of videos on Vimeo. Similarly to the DUF, the ZSM also has a temporal radius which uses frames before and after the current frame to better generate an upsampled frame. For our studies, we decreased their initial radius of 7 down to 3 due to hardware performance issues. Initially we had concerns that the ZSM pretrained model may have similar difficulties in dealing with MWIR videos like the DUF, but the results proved otherwise. When testing, we focused on the longer range videos since YOLO and ResNet struggled with detecting the targets in those videos. The results for the ZSM are shown in Table 1. It should be noted that the ZSM x2 is a bicubically downsampled version of the ZSM x4 (in our preliminary studies this was done to get quick results since the LIG is time consuming for larger images).

Table 1: Performance of detection results without SORT using videos with different resolutions.

| (a) 3500 m Vehicle 06 | | | | (b) 4000 m Vehicle 06 | | | |
|---|---|---|---|---|---|---|---|
| | P | R | F1 | | P | R | F1 |
| Original | 0.941 | 0.950 | 0.945 | Original | 0.943 | 0.943 | 0.943 |
| Bicubic x2 | 0.947 | 0.953 | 0.950 | Bicubic x2 | 0.940 | 0.940 | 0.940 |
| DUF x2 | 0.944 | 0.950 | 0.947 | DUF x2 | 0.940 | 0.937 | 0.938 |
| ZSM x2 | **0.957** | **0.957** | **0.957** | ZSM x2 | 0.960 | 0.963 | 0.962 |
| ZSM x4 | 0.950 | 0.953 | 0.952 | ZSM x4 | **0.970** | **0.973** | **0.972** |
| (c) 4500 m Vehicle 06 | | | | (d) 5000 m Vehicle 06 | | | |
| | P | R | F1 | | P | R | F1 |
| Original | 0.953 | 0.957 | 0.955 | Original | 0.928 | 0.943 | 0.936 |
| Bicubic x2 | 0.957 | 0.960 | 0.958 | Bicubic x2 | 0.928 | 0.950 | 0.939 |
| DUF x2 | **0.960** | 0.957 | 0.958 | DUF x2 | 0.928 | 0.947 | 0.937 |
| ZSM x2 | **0.960** | **0.963** | **0.962** | ZSM x2 | **0.932** | **0.953** | **0.942** |
| ZSM x4 | 0.953 | 0.957 | 0.955 | ZSM x4 | 0.922 | 0.950 | 0.936 |

It can be seen in the 3500m-5000m video results in Table 2 that the ZSM outperforms or is comparable to the best performing methods in both precision and recall rates. In the 4000 m case, which is the most difficult case in terms of achieving high performance using super-resolution images, both ZSM x2 and ZSM x4 outperforms all other cases by a slight margin. For the higher rangers like 4500m-5000m, the results were already quite good for the original, bicubic x2, and DUF x2 so there was not much room for improvement for the ZSM. But it still performs comparably to the top performing methods.

Although ZSM improves over the original resolution videos, the F1 scores are not improved by that much. So, in practical target detection applications, the value of VSR is somewhat limited. Moreover, the VSR increases the image size by two times in the ZSM x2 case, which consequently increases the computational time of the LIG detection process.

### 3.4 Customization of ZSM

We performed experiments with customizing the model by fine tuning their pretrained model with training data from the MWIR videos. The fine tuning process involves "freezing" several layers of the model. Freezing in this case means preventing the additional training data from overwriting the parameters on a set number of layers. In our case all the layers except the last 4 were frozen. This is common when fine-tuning with limited training data since the earlier layers are generally using parameters that are transferable to a variety of images or videos and the later layers are usually more specific to the training dataset used. We tested this new model with the 3500m case and found that that the model underperformed the pretrained model provided by the authors of the ZSM. Since the pretrained model was already performing so well, we decided not to experiment further with the custom training.

Table 2: Comparison of pretrained and customized ZSM models for target detection. There are 300 frames in the 3500 m video.

| | P | R | F1 |
|---|---|---|---|
| ZSM x2 | **0.930** | **0.713** | **0.807** |
| ZSM x2 Custom Model | 0.897 | 0.693 | 0.782 |

### 3.5 Further Detection Improvement Using SORT for Track Association

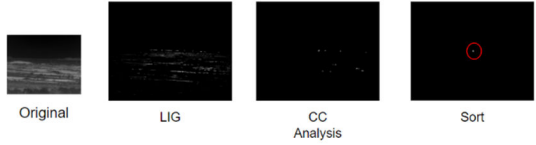

Fig. 8. Illustration of a lean workflow by combining LIG, CC, and Sort. The video super-resolution (VSR) is removed.

Table 3: Performance of target detection with and without SORT. There are 300 frames in the videos.

(a) 3500 m video

|  | P | R | F1 |
|---|---|---|---|
| Without SORT | 0.941 | 0.950 | 0.945 |
| With SORT | **1** | **0.95** | **0.974** |

(b) 4000 m video

|  | P | R | F1 |
|---|---|---|---|
| Without SORT | 0.943 | 0.943 | 0.943 |
| With SORT | **0.990** | **0.953** | **0.971** |

(c) 4500 m video

|  | P | R | F1 |
|---|---|---|---|
| Without SORT | 0.953 | 0.957 | 0.955 |
| With SORT | 0.990 | 0.943 | 0.966 |

(d) 5000 m video

|  | P | R | F1 |
|---|---|---|---|
| Without SORT | 0.928 | 0.943 | 0.936 |
| With SORT | **0.997** | **0.957** | **0.977** |

In this section, we would like to investigate the impact of SORT on the overall detection performance. Fig. 8 illustrates the complete workflow proposed by us. From Section 3.3, we observed that although the VSR using ZSM improved the detection performance, the improvement is not very significant. As a result, we decided to focus on a lean version by removing the VSR in the above pipeline. This will speed up the processing with little loss of performance.

In our experiments, the SORT algorithm was implemented after the CC Analysis step. The bounding box information of each connected component is passed along to the SORT algorithm which then attempts to correlate the bounding boxes across frames. Since the detection rate is relatively high in the earlier stages of the workflow, we believed track association will help eliminate the majority of false positives. Our initial testing, presented in Table 3, has confirmed our hypothesis. On the 300 frames in the 3500m video, the use of the SORT algorithm eliminated all the false positives. The results in other ranges also improved quite a lot. The SORT algorithm can help further distinguish between the background anomalies and the actual target. It should be noted that not all frames have a detection due to dim targets in some frames.

### 3.6 Computational complexity

The proposed framework is slow due to the use of LIG, which is the bottleneck. Even with parallel implementation for LIG using PARFOR in Matlab, it took approximately 75 seconds to process one frame, as can be seen in Table 4.

Table 4: Computational times for the proposed target detection framework.

| Method | Language | Time for 300 Frames (s) |
|---|---|---|
| LIG | MATLAB | 63000 |
| LIG Parallel | MATLAB | **22000** |

### 3.7 Subjective Evaluation

Results for the workflow with and without SORT are included in Fig. *9* to Fig. *12* for comparison purposes. It can be seen that, in certain frames, SORT helps to eliminate the false positives that are detected by LIG.

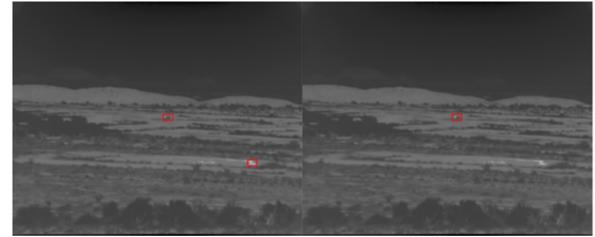

(a) Without SORT  (b) With SORT

Fig. 9. 3500 m frame comparing with and without SORT.

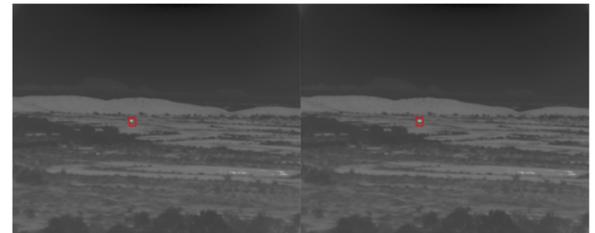

(a) Without SORT  (b) With SORT

Fig. 10. 4000 m frame comparing with and without SORT.

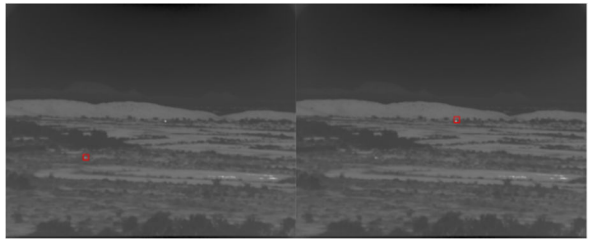

(a) Without SORT    (b) With SORT

Fig. 11. 4500 m frame comparing with and without SORT.

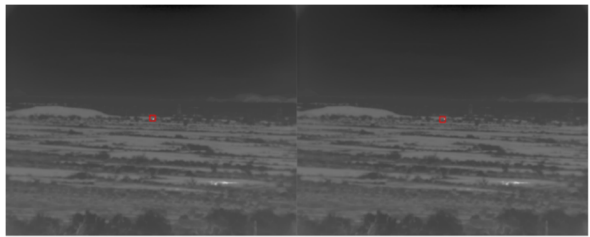

(a) Without SORT    (b) With SORT

Fig. 12. 5000 m frame comparing with and without SORT.

## 4 Conclusions

In this research, we focus on small target detection in long range infrared videos. A flexible and modular framework that contains a video super-resolution module, an unsupervised target detector, a connected component analysis module, and a track association module, was developed. The LIG method 5] was modified, sped up, and applied to four videos that were collected from 3500 m to 5000 m. Three video super-resolution algorithms were compared and one algorithm was found to improve the overall detection performance. However, the video super-resolution algorithm (ZSM) only improved the detection performance slightly. It was observed that the LIG algorithm worked quite well for target detection even at long ranges. Finally, the most dramatic improvement occurred when a simple target association algorithm known as SORT was incorporated into our framework.

One future direction is to further improve the computational complexity of the framework. The bottleneck is the LIG algorithm. Fortunately, the LIG uses a sliding window to scan through the whole image, implying that a parallel processing based on GPU may be a good research direction.

**Acknowledgements** This work was partially supported by US government PPP program. The views, opinions and/or findings expressed are those of the author(s) and should not be interpreted as representing the official views or the U.S. Government.

## References

1. Chen, C. L. P., Li, H., Wei, Y., Xia, T., Tang, Y. Y.: A local contrast method for small infrared target detection. *IEEE Trans. Geosci. Remote Sens.*, vol. 52, no. 1, pp. 574–581. (2014)
2. Wei, Y., You, X., Li, H.: Multiscale patch-based contrast measure for small infrared target detection. *Pattern Recognit.*, vol. 58, pp. 216–226. (2016)
3. Gao, C., Meng, D., Yang, Y., Wang, Y., Zhou, X., Hauptmann, A. G.: Infrared patch-image model for small target detection in a single image. *IEEE Trans. Image Process.*, vol. 22, no. 12, pp. 4996–5009. (2013)
4. Zhang, L., Peng, L., Zhang, T., Cao, S., Peng, Z.: Infrared small target detection via non-convex rank approximation minimization joint l2,1 norm. *Remote Sens.*, vol. 10, no. 11, pp. 1–26. (2018)
5. Zhang, H., Zhang, L., Yuan, D., Chen, H.: Infrared small target detection based on local intensity and gradient properties. *Infrared Phys. Technol.*, vol. 89, pp. 88–96. (2018)
6. Chen, Y., Zhang, G., Ma, Y., Kang, J. U., Kwan, C.: Small Infrared Target Detection based on Fast Adaptive Masking and Scaling with Iterative Segmentation. IEEE Geoscience and Remote Sensing Letters, submitted, 8/11/2020.
7. Kwan, C., Chou, B., Kwan, L. M.: A Comparative Study of Conventional and Deep Learning Target Tracking Algorithms for Low Quality Videos. 15th International Symposium on Neural Networks. (2018)
8. Demir, H. S., Cetin, A. E.: Co-difference based object tracking algorithm for infrared videos. IEEE International Conference on Image Processing (ICIP), Phoenix, AZ, pp. 434-438. (2016)
9. Kwan, C., Chou, B., Yang, J., Tran, T.: Compressive object tracking and classification using deep learning for infrared videos. Proc. SPIE 10995, *Pattern Recognition and Tracking (Conference SI120)*. (2019)
10. Kwan, C., Chou, B., Yang, J., Tran, T.: Target tracking and classification directly in compressive measurement for low quality videos. SPIE 10995, Pattern Recognition and Tracking XXX, 1099505. 13 May (2019)
11. Kwan, C., Chou, B., Echavarren, A., Budavari, B., Li, J., Tran, T.: Compressive vehicle tracking using deep learning. *IEEE Ubiquitous Computing, Electronics & Mobile Communication Conference,* New York City. (2018)
12. Kwan, C., Gribben, D., Tran, T.: Multiple Human Objects Tracking and Classification Directly in Compressive Measurement Domain for Long Range Infrared Videos. *IEEE Ubiquitous Computing, Electronics & Mobile Communication Conference*, New York City. (2019)
13. Kwan, C., Gribben, D., Tran, T.: Tracking and Classification of Multiple Human Objects Directly in Compressive Measurement Domain for Low Quality Optical Videos. *IEEE Ubiquitous Computing, Electronics & Mobile Communication Conference*. New York City. (2019)

14. Kwan, C., Chou, B., Yang, J., Tran, T.: Deep Learning based Target Tracking and Classification Directly in Compressive Measurement for Low Quality Videos. *Signal & Image Processing: An International Journal (SIPIJ)*. November 16. (2019)
15. Kwan, C., Chou, B., Yang, J., Rangamani, A., Tran, T., Zhang, J., Etienne-Cummings, R.: Target tracking and classification directly using compressive sensing camera for SWIR videos. *Journal of Signal, Image, and Video Processing*. June 7. (2019)
16. Kwan, C.; Chou, B.; Yang, J.; Rangamani, A.; Tran, T.; Zhang, J.; Etienne-Cummings, R.: Target tracking and classification using compressive measurements of MWIR and LWIR coded aperture cameras. *Journal Signal and Information Processing*. **2019**, *10*, 73–95.
17. Kwan, C., Gribben, D., Rangamani, A., Tran, T., Zhang, J., Etienne-Cummings, R.: Detection and Confirmation of Multiple Human Targets Using Pixel-Wise Code Aperture Measurements. *J. Imaging*. 6(6), 40. (2020)
18. Kwan, C., Chou, B., Yang, J., Tran, T.: Deep Learning based Target Tracking and Classification for Infrared Videos Using Compressive Measurements. *Journal Signal and Information Processing*. November. (2019)
19. Kwan, C., Chou, B., Yang, J., Rangamani, A., Tran, T., Zhang, J., Etienne-Cummings, R.: Deep Learning based Target Tracking and Classification for Low Quality Videos Using Coded Aperture Camera. *Sensors*, 19(17), 3702, August 26. (2019)
20. Lohit, S., Kulkarni, K., Turaga, P. K.: Direct inference on compressive measurements using convolutional neural networks. *Int. Conference on Image Processing*. 1913-1917. (2016)
21. Adler, A., Elad, M., Zibulevsky, M.: Compressed Learning: A Deep Neural Network Approach. arXiv:1610.09615v1 [cs.CV]. (2016)
22. Xu, Y., Kelly, K. F.: Compressed domain image classification using a multi-rate neural network," arXiv:1901.09983 [cs.CV]. (2019)
23. Wang, Z. W., Vineet, V., Pittaluga, F., Sinha, S. N., Cossairt, O., Kang, S. B.: Privacy-Preserving Action Recognition Using Coded Aperture Videos. *IEEE Conference on Computer Vision and Pattern Recognition (CVPR) Workshops*. (2019)
24. Vargas, H., Fonseca, Y., Arguello, H.: Object Detection on Compressive Measurements using Correlation Filters and Sparse Representation. *26th European Signal Processing Conference (EUSIPCO)*. 1960-1964. (2018)
25. Değerli, A., Aslan, S., Yamac, M., Sankur, B., Gabbouj, M.: Compressively Sensed Image Recognition. *7th European Workshop on Visual Information Processing (EUVIP)*, Tampere. (2018)
26. Latorre-Carmona, P., Traver, V. J., Sánchez, J. S., Tajahuerce, E.: Online reconstruction-free single-pixel image classification. *Image and Vision Computing*. (2018)
27. Kwan, C., Gribben, D., Chou, B., Budavari, B., Larkin, J., Rangamani, A., Tran, T., Zhang, J., Etienne-Cummings, R.: Real-Time and Deep Learning based Vehicle Detection and Classification using Pixel-Wise Code Exposure Measurements. *Electronics*, June 18. (2020)
28. Jo, Y., Oh, S. W., Kang, J., Kim, S. J.: Deep video super-resolution network using dynamic upsampling filters without explicit motion compensation. In *Proceedings of the IEEE conference on computer vision and pattern recognition* (pp. 3224-3232). (2018)
29. Xiang, X.: Mukosame/Zooming-Slow-Mo-CVPR-2020. GitHub, 2020, github.com/Mukosame/Zooming-Slow-Mo-CVPR-2020
30. Xiang, X., Tian, Y., Zhang, Y., Fu, Y., Allebach, J. P., Xu, C.: Zooming Slow-Mo: Fast and Accurate One-Stage Space-Time Video Super-Resolution. In *Proceedings of the IEEE/CVF Conference on Computer Vision and Pattern Recognition* (pp. 3370-3379). (2020)
31. Bewley, A., Ge, Z., Ott, L., Ramos, F., Upcroft, B.: Simple Online and Realtime Tracking. arXiv 2016. arXiv preprint arXiv:1602.00763.
32. SENSIAC dataset, https://www.news.gatech.edu/2006/03/06/sensiac-center-helps-advance-military-sensing. Accessed November 9, 2020.